%% file: main.tex
\documentclass{article}
\usepackage{iclr2023_conference,times}

\input{math_commands.tex}

\usepackage{hyperref}
\usepackage{url}

\usepackage{xcolor} 
\usepackage{booktabs}       
\usepackage{amsfonts}       
\usepackage{nicefrac}       
\usepackage{microtype}      
\usepackage{multirow}
\usepackage{graphicx}
\usepackage{wrapfig}
\usepackage{amssymb}

\title{Accelerating Neural Self-Improvement\\ via Bootstrapping}

\author{Kazuki Irie$^{1}$ ~ J\"urgen Schmidhuber$^{1,2}$\\
  $^1$The Swiss AI Lab, IDSIA, USI \& SUPSI, Lugano, Switzerland \\
  $^2$AI Initiative, KAUST, Thuwal, Saudi Arabia \\
  \texttt{\{kazuki, juergen\}@idsia.ch}
}

\iclrfinalcopy
\begin{document}

\maketitle

\begin{abstract}
Few-shot learning with sequence-processing neural networks (NNs) has recently attracted a new wave of attention in the context of large language models.
In the standard $N$-way $K$-shot learning setting,
an NN is explicitly optimised to learn to classify unlabelled inputs by observing a sequence of $NK$ labelled examples.
This pressures the NN to learn a learning algorithm that achieves optimal performance, given the limited number of training examples. Here we study an auxiliary loss that encourages \textit{further} acceleration of few-shot learning,
by applying recently proposed bootstrapped meta-learning to NN few-shot learners:
we optimise the $K$-shot learner to match its own performance achievable by observing \textit{more} than $NK$ examples, using \textit{only} $NK$ examples.
Promising results are obtained on the standard Mini-ImageNet dataset.
Our code is public.\footnote{\url{https://github.com/IDSIA/modern-srwm}}
\end{abstract}

\section{Introduction \& Motivation}
\label{sec:intro}
Since the seminal works by \citet{hochreiter2001learning, younger1999fixed, cotter1991learning, cotter1990fixed},
many sequence processing neural networks (NNs)---including recurrent NNs (RNNs) e.g., in  \citet{SantoroBBWL16} but also Transformer variants \citep{trafo} as in \citet{mishra2018a}---have been trained as 
a meta-learner \citep{schmidhuber1987evolutionary, Schmidhuber:92selfref}
that learns to learn to solve a task by observing sequences of training examples (i.e., pairs of inputs and their labels).
A popular application of such an approach is to build a sample efficient \textit{few-shot learner} capable of adaption to a new (but related) task from few training examples or demonstrations \citep{FinnAL17}, which typically achieves respectable performance in the standard benchmarks for few-shot learning (e.g., image classification), see e.g., \citet{mishra2018a, munkhdalai2017meta}. 
More recently, such ``on-the-fly'' learning capability of sequence processing NNs has attracted broader interests in the context of large language models (LLMs; \citet{gpt2}).
In fact, the task of language modelling itself has a form of \textit{sequence processing with error feedback} (essential for meta-learning \citep{schmidhuber1990making}): the correct label to be predicted is fed to the model with a delay of one time step in an auto-regressive manner.
Trained on a large amount of text covering a wide variety of credit assignment paths, LLMs exhibit certain sequential few-shot learning capabilities in practice \citep{gpt3}.
This was rebranded as \textit{in-context learning}, and
has been the subject of numerous recent studies (e.g., \citet{XieRL022, chan2022data, chan2022transformers, kirsch2022general, akyurek2022learning, von2022transformers, dai2022can}).\looseness=-1

Here our focus is the standard \textit{sequential few-shot classification}.
Let $d$, $N$, $K$, $L$ be positive integers.
In sequential $N$-way $K$-shot classification settings, a sequence processing NN with a parameter set $\theta \in \mathbb{R}^L$ observes a pair ($\vx_t$, $y_t$) where $\vx_t \in \mathbb{R}^{d}$ is the input and $y_t \in \{1,..., N\}$ is its label at each step $t \in \{1, ..., N\cdot K\}$, corresponding to $K$ examples for each one of $N$ classes.
After the presentation of these $N\cdot K$ examples (often called the \textit{support set}), one extra input $\vx \in \mathbb{R}^{d}$ (often called the \textit{query}) is fed to the model without its label---in practice, an unknown label token $\varnothing$ is fed instead, e.g., $\varnothing=0$---and the task of the model is to predict its label.
During training, the parameters of the model $\theta$ are optimised to maximise the probability $p(y|(\vx_1,y_1), ..., (\vx_{N\cdot K},y_{N\cdot K}), (\vx, \varnothing);\theta)$ of the correct label $y \in \{1,..., N\}$ of the input query $\vx$.
Since class-to-label associations are randomised and unique to each sequence ($(\vx_1,y_1), ..., (\vx_{N\cdot K},y_{N\cdot K}), (\vx, \varnothing)$), each such a sequence can be used as a new (few-shot) learning experience to train the model.

While $K$ is a pre-defined constant in the setting above,
 ideally, we want to obtain a learning algorithm (encoded by the NN with parameters $\theta$) that achieves the ``best possible performance'' with the minimum number of few-shot learning examples i.e., here $N \cdot K$.
We disregard mathematical rigour for now, in favour of an intuitive description that illustrates the motivation of our work (Sec.~\ref{sec:method}).
We informally assume that there exists a convenient error function $\mathcal{L}: (\mathbb{R}^L, \mathbb{N}) \rightarrow \mathbb{R}$ that, given a number of learning steps $t \in \mathbb{N}$, evaluates the performance $\mathcal{L}(\theta, t)$ (e.g., the lower the better) of $\theta \in \mathbb{R}^L$ as a learning algorithm on a certain task which we assume to be fixed.
We also assume a certain performance value $\mathcal{L}^* \in \mathbb{R}$ which we consider as ``optimal'' for this task.
We further assume that there exists another convenient function $\tau: (\mathbb{R}^L, \mathbb{R}) \rightarrow \mathbb{N}$ that tells us the number of learning steps $\tau(\theta, \mathcal{L}^*)$ required by the learning algorithm parameterised by $\theta$ to achieve the performance $\mathcal{L}^*$.
Then, the learning algorithm we want to find is the solution of:
\begin{align}
\mathop\text{minimise} \limits_{\theta} \tau(\theta, \mathcal{L}^*)
\end{align}
In general, we do not have access to such a function $\tau$.
But if we assume that, given $\theta$, performance is monotonic w.r.t.~the number of learning steps, i.e., for $t_1, t_2 \in \mathbb{N}$ such that $t_1 < t_2$, we have $\mathcal{L}(\theta, t_2) \leq \mathcal{L}(\theta, t_1)$, i.e., $\mathcal{L}(\theta, t_2)$ is better than $\mathcal{L}(\theta, t_1)$, 
and if we further assume that we have access to the ``models'' $\mW^{\theta}_{t_1}$ and  $\mW^{\theta}_{t_2}$ trained on the fly by the learning algorithm $\theta$ at the corresponding steps $t_1$ and $t_2$,
there are several ways of explicitly optimising $\theta$
such that the performance of $\mW^{\theta}_{t_1}$ ``matches'' that of $\mW^{\theta}_{t_2}$, e.g., via knowledge distillation.
Such an auxiliary task encourages $\theta$ to achieve performance $\mathcal{L}(\theta, t_2)$ within the number of steps $t_1$ which is less than $t_2$ required by the current $\theta$.

This idea directly relates to that of \textit{bootstrapped meta-learning} by \citet{FlennerhagSZHS022} but applied to few-shot learning with sequence processing NNs.
Here we explore this approach in self-modifying NNs based on a modern variant \citep{IrieSCS22} of self-referential weight matrices \citep{Schmidhuber:92selfref, Schmidhuber:93selfreficann} that has several appealing properties including its close connection to Transformers (Sec.~\ref{sec:model}).

\section{Method}
\label{sec:method}
The main idea of this work is to apply \textit{bootstrapped training} \citep{FlennerhagSZHS022} to improve few-shot learning with sequence processing NNs.
In what follows, we specify the details of our sequence processor (Sec.~\ref{sec:model}), and describe the bootstrapped training objective (Sec.~\ref{sec:bootstrap}).

\subsection{Background: Linear Transformers and Self-Referential Weight Matrices}
\label{sec:model}
\begin{figure}[t]
    \begin{center}
        \includegraphics[width=0.95\columnwidth]{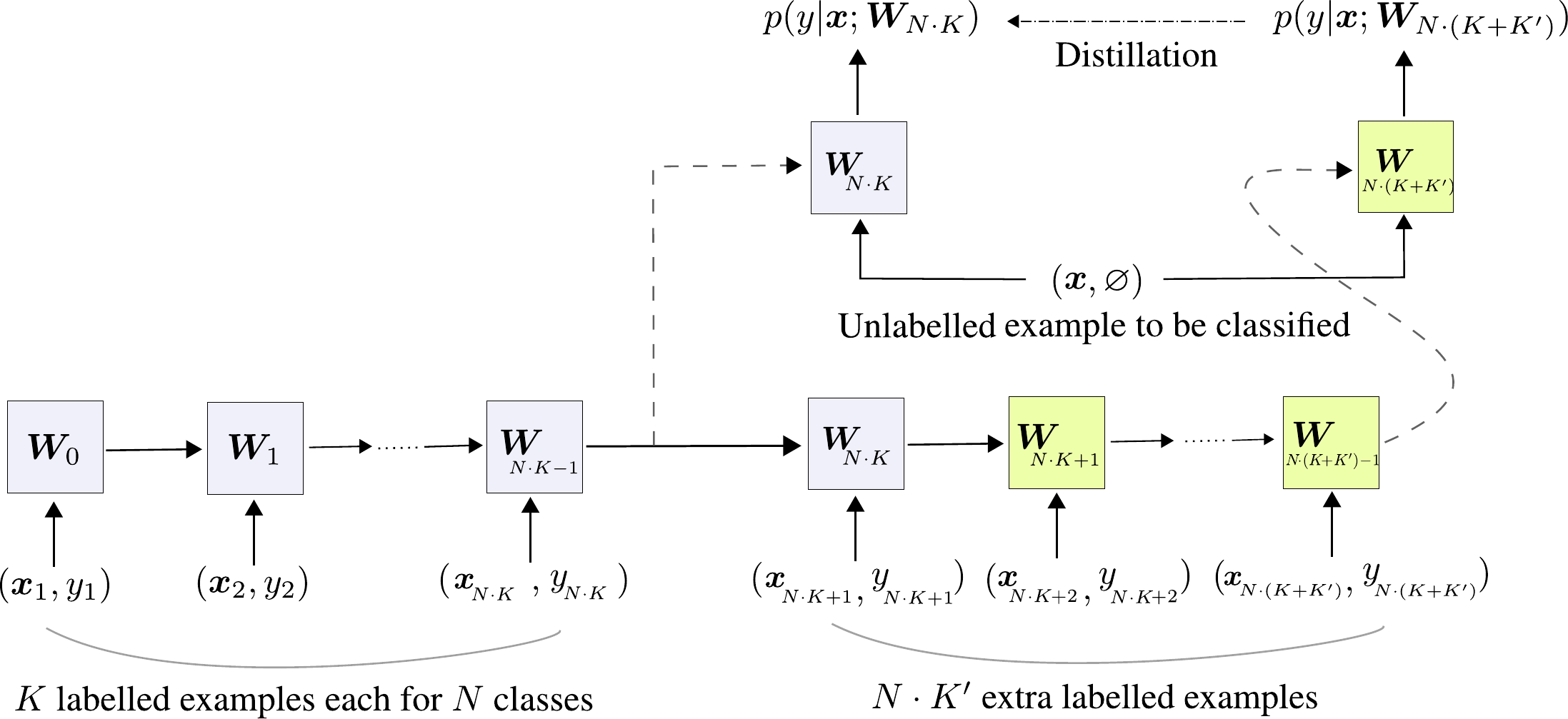}
        \caption{An illustration of bootstrapped training of a self-modifying WM $\mW_0$.
        In $N$-way $K$-shot learning (in \textit{grey}), $\mW_0$ observes a sequence of $N \cdot K$ labelled examples, and yields $\mW_{N \cdot K}$ which is used to classify an unlabelled input $\vx$.
        In bootstrapped training (in \textit{light green}), we feed $N \cdot K'$ extra inputs to $\mW_{N \cdot K}$ to obtain $\mW_{N \cdot (K+K')}$.
        We evaluate $\mW_{N \cdot (K+K')}$ using the same test example $\vx$, and its output is used as a teaching signal to train $\mW_{N \cdot K}$ via knowledge distillation.}
        \label{fig:bootstrap_sm}
    \end{center}
\end{figure}

\textit{Bootstrapped training} (Sec.~\ref{sec:bootstrap}) can be applied to any sequence-processing NNs trained for few-shot learning.
An application to NNs with a self-modifying weight matrix (WM) yields a particularly intuitive setting (illustrated in Figure \ref{fig:bootstrap_sm}; commented on later).
Here we use the \textit{modern self-referential weight matrix} (SRWM; \citet{IrieSCS22}) as a generic self-modifying WM.
An SRWM is a WM that sequentially modifies itself as a response to a stream of input observations \citep{Schmidhuber:92selfref, Schmidhuber:93selfreficann}.
The modern SRWM belongs to the family of linear Transformers a.k.a.~Fast Weight Programmers (FWPs; \citet{Schmidhuber:91fastweights, katharopoulos2020transformers, choromanski2020rethinking, peng2021random, schlag2021linear, irie2021going}).
Linear Transformers and FWPs are an important class of the now popular Transformers \citep{trafo}:
unlike the standard Transformers whose state size linearly grows with the context length, the state size of FWPs is constant w.r.t.~sequence length (like in the standard RNNs).
This is an attractive property, since several open problems with Transformers relate to its explicitly limited context window size (see e.g., \citet{laskin2022context, reed2022generalist}).
Such a property is important also because potential extensions of the current in-context learning setting to further \textit{in-context continual learning} etc.~would require handling context lengths much larger than those supported by today's Transformers.
Moreover, the duality between linear attention and FWPs \citep{schlag2021linear}---and likewise, between linear attention and linear layers trained by the gradient descent learning algorithm \citep{irie2022dual, aizerman1964theoretical}---have played a key role in certain theoretical analyses of few-shot learning capabilities of Transformers \citep{von2022transformers, dai2022can}.

The dynamics of an SRWM \citep{IrieSCS22} are as follows. Let $d_\text{in}$, $d_\text{out}$, $t$ be positive integers, and $\otimes$ denote outer product. At each time step $t$, an SRWM $\mW_{t-1} \in \mathbb{R}^{(d_\text{out} + 2 * d_\text{in} + 1) \times d_\text{in}}$ observes an input $\vx_t \in \mathbb{R}^{d_\text{in}}$, and outputs $\vy_t \in \mathbb{R}^{d_\text{out}}$, while also updating itself to $\mW_{t}$ as follows:
\begin{align}
\label{eq:srm_start}
\vy_t, \vk_t, \vq_t, \beta_t &= \mW_{t-1} \vx_t \\
\label{eq:srm_key}
\vv_t = \mW_{t-1} \phi(\vq_t)
&; \, \bar{\vv}_t = \mW_{t-1} \phi(\vk_t) \\
\label{eq:srm_end}
\mW_{t} &= \mW_{t-1} + \sigma(\beta_t)(\vv_t - \bar{\vv}_t) \otimes \phi(\vk_t)
\end{align}
where $\vv_t, \bar{\vv}_t \in \mathbb{R}^{(d_\text{out} + 2 * d_\text{in} + 1)}$ are value vectors, $\vq_t \in \mathbb{R}^{d_\text{in}}$
 and $\vk_t \in \mathbb{R}^{d_\text{in}}$ are query and key vectors,
 and $\beta_t \in \mathbb{R}$ is the learning rate.
 $\sigma$ and $\phi$ denote sigmoid and softmax functions respectively.
  $\phi$ is typically also applied to $\vx_t$ in Eq.~\ref{eq:srm_start}, but not in the experiments presented here (following \citet{IrieSCS22}'s few-shot image classification setting).
$\mW_0$ is the only trainable parameters of this layer, that encodes the initial self-modification algorithm.
In practice, we use the layer above as a replacement to the self-attention layer in the Transformer architecture \citep{trafo} with all other components such as the two-layer feedforward block, and we also use the multi-head version of the SRWM computation above.

\subsection{Bootstrapped Training}
\label{sec:bootstrap}
\textit{Bootstrapped training} of few-shot learning NNs is directly inspired by \citet{FlennerhagSZHS022} (and also relates to \citet{tack2022meta}).
The overall process is depicted in Figure \ref{fig:bootstrap_sm}, and works as follows.

Let $N$, $K$, $K'$, $d$ be positive integers.
In $N$-way $K$-shot classification (see Sec.~\ref{sec:intro}), an SRWM (see Sec.~\ref{sec:model}) $\mW_0$ observes a sequence of $N \cdot K$ labelled examples, and the resulting WM, $\mW_{N \cdot K}$, is used to classify an unlabelled query input $\vx \in \mathbb{R}^d$.
$\mW_0$ is trained to produce a sequence of self-modifications that result in $\mW_{N \cdot K}$ capable of such classification.
\textit{Bootstrapped training} augments the setting above by continuing to feed $N \cdot K'$ extra labelled inputs to $\mW_{N \cdot K}$ to obtain $\mW_{N \cdot (K + K')}$.
Then, we feed the same unlabelled example $\vx$ to $\mW_{N \cdot (K + K')}$, and use its output distribution as a teaching signal to train $\mW_{N \cdot K}$ via knowledge distillation (\citet{ba14, hinton2014}; which \citet{Schmidhuber:91chunker} calls \textit{collapsing}).
For this to work, we assume $\mW_{N \cdot (K + K')}$ to perform better than $\mW_{N \cdot K}$ thanks to extra learning examples (which is effectively the case in standard few-shot learning scenarios).
By denoting the output distribution of the model with weights $\mW_{N\cdot K}$ as $\vp_{N\cdot K}(\cdot|\vx) = \vp(\cdot |\vx; \mW_{N\cdot K})$, the training objective function is:
\begin{align}
\label{eq:loss}
\beta_1 \CE(\vq, \vp_{N\cdot K}; \vx) + \beta_2 \CE(\sg(\vp_{N\cdot (K+K')}), \vp_{N\cdot K}; \vx) + \beta_3 \CE(\vq, \vp_{N\cdot (K+K')}; \vx)
\end{align}
where $\beta_1, \beta_2, \beta_3 \in \mathbb{R}_{>0}$ are scaling factors,
$\CE$ denotes the standard cross-entropy loss:
\begin{align}
\label{eq:ce}
\CE(\vq, \vp; \vx) &= - \sum_{1 \leq y \leq N} \vq(y| \vx) \log(\vp(y| \vx))
\end{align}
and $\vq(\cdot|\vx) \in \{0, 1\}^N$ denotes the one-hot vector encoding the correct label of $\vx \in \mathbb{R}^d$.
In Eq.~\ref{eq:loss}, the first and third terms correspond to the standard $N$-way $K$-shot and $(K+K')$-shot learning losses respectively, while the second term is the distillation loss (where $\sg$ denotes the ``stop-gradient'' operation; we want $\mW_{N \cdot K}$ to match the performance of $\mW_{N \cdot (K+K')}$, but not the opposite).

\begin{table*}[t]
\caption{Few-shot image classification accuracies (\%) on Mini-ImageNet with bootstrapped training using $K' \cdot N$ extra examples for various values of $K'$.
The row ``$K'=$ None'' corresponds to the pure $N$-way $K$-shot learning without any auxiliary loss, i.e., $\beta_1=1$, $\beta_2=0$ and $\beta_3=0$ in Eq.~\ref{eq:loss}.
}
\label{tab:few_shot}
\vskip 0.15in
\begin{center}
\begin{tabular}{rcc}
\toprule
  $K'$  & \multicolumn{1}{c}{$K=5$, $K_{\text{test}}=5$} & \multicolumn{1}{c}{$K=10$, $K_{\text{test}}=10$} \\ \midrule
None & 60.6 $\pm$ 0.4 &  64.1 $\pm$ 0.4  \\
1 & 61.4 $\pm$ 0.5 & 66.9 $\pm$ 0.3  \\
5  & 62.6 $\pm$ 0.3 & \textbf{68.3} $\pm$ 0.3  \\
10 &  \textbf{63.3} $\pm$ 0.4 & 67.6 $\pm$ 0.2  \\
\bottomrule
\end{tabular}
\end{center}
\vskip -0.1in
\end{table*}

\begin{table*}[t]
\caption{Few-shot image classification accuracies (\%) on Mini-ImageNet of models trained with or without bootstrapped training (``Bootstrap'' Yes/No) for various $K_{\text{test}}$. $\beta_1=1$ and  $\beta_3=1$ in all cases. If bootstrapping is used, $\beta_2=10.0$ for the $K=10$ case, and $\beta_2=5.0$ for the $K=5$ case.
}
\label{tab:various_shot}
\vskip 0.15in
\begin{center}
\begin{tabular}{rcccc}
\toprule
   & \multicolumn{2}{c}{$K=5$, $K'=10$} & \multicolumn{2}{c}{$K=10$, $K'=5$} \\  \cmidrule(r){2-3} \cmidrule(r){4-5} 
Bootstrap   & No & Yes & No & Yes \\ \midrule
$K_{\text{test}}$ = 1  & 44.8 $\pm$ 0.2 &  \textbf{45.9} $\pm$ 0.5 & 38.7 $\pm$ 0.2 &  44.0 $\pm$ 0.3 \\
5  &  60.1 $\pm$ 0.4 & 63.3 $\pm$ 0.4 & 59.0 $\pm$ 0.4  & \textbf{63.7} $\pm$ 0.2  \\
10 & 64.0 $\pm$ 0.4 &  66.6 $\pm$ 0.4 & 64.6 $\pm$ 0.4   & \textbf{68.3} $\pm$ 0.3   \\
15 & 65.2 $\pm$ 0.4 & 67.3 $\pm$ 0.4 & 66.2 $\pm$ 0.2 & \textbf{69.5}  $\pm$ 0.4 \\
\bottomrule
\end{tabular}
\end{center}
\vskip -0.1in
\end{table*}

\section{Experiments}
\label{sec:exp}
\textbf{Experimental Setting.}
We conduct experiments on the standard Mini-ImageNet dataset \citep{VinyalsBLKW16, RaviL17} for few-shot image classification.
All our experimental setting follows the one of \citet{IrieSCS22},
except that we use a single architecture for all $K$, and we report results with the standard deviation instead of the commonly used \citep{RaviL17} 95\% confidence interval (which is below 0.1 in all our cases).
All our models have 3 SRWM layers, with 16 heads and a total hidden dimension of 256 in each layer, and the 2-layer feedforward block has an inner dimension of 2048. We train them with a batch size of 16 using the warm-up learning rate scheduling as in \citet{IrieSCS22}.
We refer to \citet{IrieSCS22} and our public code for further experimental details.

\textbf{Main Results.}
We conduct experiments on the standard Mini-ImageNet dataset \citep{VinyalsBLKW16, RaviL17} for few-shot image classification.
Our base architecture is similar to the one of \citet{IrieSCS22} (but unlike these authors we use a single architecture for all $K$).
We refer to \citet{IrieSCS22} for all the basic experimental details.
We first evaluate various values of $K'$ (number of extra ``shots'' used for bootstrapping) in the 5-way 5-shot and 10-shot settings.
Table \ref{tab:few_shot} shows the results.
Here models are tested in the $K_{\text{test}}$-shot setting where $K_{\text{test}}=K$ (as in training).
In all cases, we observe improvements by bootstrapped training over the baselines without any auxiliary losses (slight improvements in the 5-shot case).
Now in Table \ref{tab:various_shot}, we fix $K$ and $K'$, and show how bootstrapped training affects the performance of the model evaluated with different numbers of examples at test time (i.e., for different $K_{\text{test}}$).
Here, bootstrapping is disabled in the baselines (``Bootstrap No'') but the auxiliary $(K+K')$-shot loss is used ($\beta_1 = \beta_3 = 1$) for fair comparison.
We observe that bootstrapping not only improves $K_{\textit{test}}$-shot learning where $K_{\textit{test}}=K$ (i.e., where the auxiliary bootstrapping loss is applied in training), it also yields a chain of improvements at any $K_{\textit{test}}$, resulting in an overall acceleration of few-shot learning.

\textbf{Remarks \& Current Limitations.}
In Table \ref{tab:various_shot}, we observe that the model trained for $K=10$ underperforms the one trained with $K=5$ on 1-shot learning task ($K_{\text{test}}=1$)---in fact, both of them underperform the model specifically tuned for 1-shot learning ($\approx 47\%$ accuracy is reported in \citet{IrieSCS22}).
This trend may change by training the model in the ``delayed label'' setting (where correct labels are fed to the model one step later; see, \citet{IrieSCS22}).
Also, we have not managed to obtain any improvement in the $1$-shot learning case (i.e., the same experiments as in Table \ref{tab:few_shot} but with $K=1$):
we speculate that this case is difficult, since with $N=5$ and $K=1$, our SRWM can only generate rank-5 updates.
Optimally tuning the model for 1-shot learning may leave little leeway for further improvements.
Also importantly, so far we have not succeeded at confirming the improvements on another dataset, tiered-ImageNet \citep{RenTRSSTLZ18}:
we ran the configuration used for Mini-ImageNet (without changing any hyper-parameters, including those for the model architecture), without observing out-of-the-box improvements.
Further empirical investigations are still needed to determine under which conditions  exactly bootstrapped training can help.
In fact, this is an open question for knowledge distillation in general.

\section{Conclusion}
\label{sec:diss}
We study bootstrapped training of sequence-processing neural networks (NNs) for few-shot learning (FSL).
Bootstrapped training encourages an FSL NN to accelerate FSL by letting it chase its own performance achievable with {\em more} training examples, using {\em less} examples.
Our preliminary experiments on the standard Mini-ImageNet yield promising results.
Further investigations using other datasets and tasks, e.g, imitation learning \citep{xu2022hyper}, are needed to confirm these trends.\looseness=-1

\section*{Acknowledgements}
This research was partially funded by ERC Advanced grant no: 742870, project AlgoRNN,
and by Swiss National Science Foundation grant no: 200021\_192356, project NEUSYM.
We are thankful for hardware donations from NVIDIA and IBM.
The resources used for this work were partially provided by Swiss National Supercomputing Centre (CSCS) project s1145 and s1154.

\bibliography{references}
\bibliographystyle{iclr2023_conference}

\end{document}

%% file: math_commands.tex
\usepackage{amsmath,amsfonts,bm}

\def\eqref#1{equation~\ref{#1}}

\def\1{\bm{1}}

\def\vk{{\bm{k}}}

\def\vp{{\bm{p}}}
\def\vq{{\bm{q}}}

\def\vv{{\bm{v}}}

\def\vx{{\bm{x}}}
\def\vy{{\bm{y}}}

\def\mW{{\bm{W}}}

\DeclareMathAlphabet{\mathsfit}{\encodingdefault}{\sfdefault}{m}{sl}
\SetMathAlphabet{\mathsfit}{bold}{\encodingdefault}{\sfdefault}{bx}{n}

\newcommand{\sg}{\mathrm{sg}}
\newcommand{\CE}{\mathrm{CE}}

